\documentclass[10pt, a4paper]{article}
\usepackage{lrec2022} % this is the new LREC2022 Style
\usepackage{multibib}
\usepackage{graphicx}
\usepackage{tabularx}
\usepackage{soul}
\usepackage{titlesec}
\titleformat{\section}{\normalfont\large\bfseries\center}{\thesection.}{1em}{}
\titleformat{\subsection}{\normalfont\SmallTitleFont\bfseries\raggedright}{\thesubsection.}{1em}{}
\titleformat{\subsubsection}{\normalfont\normalsize\bfseries\raggedright}{\thesubsubsection.}{1em}{}
\renewcommand\thesection{\arabic{section}}
\renewcommand\thesubsection{\thesection.\arabic{subsection}}
\renewcommand\thesubsubsection{\thesubsection.\arabic{subsubsection}}
\usepackage{epstopdf}
\usepackage[utf8]{inputenc}

\usepackage{hyperref}
\usepackage{xstring}

\usepackage{color}
\usepackage{appendix}

\title{Dataflow Dialogue Generation}

\name{Joram Meron, Victor Guimarães } 

\address{Telepathy Labs GmbH \\
         36 Militärstrasse, Zurich, Switzerland  \\
         \{joram.meron, victor.guimaraes\}@telepathy.ai\\}

\abstract{
We demonstrate task-oriented dialogue generation within the dataflow dialogue paradigm. We show an example of agenda driven dialogue generation for the MultiWOZ domain, and an example of generation without an agenda for the SMCalFlow domain, where we show an improvement in the accuracy of the translation of user requests to dataflow expressions when the generated dialogues are used to augment the translation training dataset.
\\ 
\newline 
 }

\begin{document}

\maketitleabstract

\section{Introduction}

%Task oriented dialogue systems are ubiquitous, allowing organizations to provide various services to their clients with low cost and high availability.
%
%Until the challenges facing the use of large language models (and other end-to-end systems) for these tasks are solved (uncontrolled behaviours such as hallucinations and untruthfulness), dialogue systems in actual production are still manually created for the specific applications - either manually generated rules, and/or manually curated data.

\cite{andreas2020task-oriented} has introduced the use of the dataflow (DF) paradigm to dialogue modelling, using computational graphs to hierarchically represent user requests, data, and the dialogue history. The approach was demonstrated in SMCalFlow, a large dialogue dataset, labelled with dataflow annotations.

In a follow-up work, OpenDF\footnote{https://github.com/telepathylabsai/OpenDF}, an implementation of a dataflow dialogue framework, was presented in \cite{simplifying}, which also suggested a simplified version of the SMCalFlow annotations.

Using the OpenDF system, a dataflow  implementation of MultiWOZ \cite{budzianowski-etal-2018-multiwoz} was presented in \cite{mwoz_df}, demonstrating some of the benefits of the dataflow dialogue paradigm.
 
This work takes an additional step by adding a user simulator component to the dataflow dialogue system, resulting in a dialogue generator, which can generate both sides (user and agent) of a complete conversation.

%In this work, we build upon the dataflow implementation of the MultiWOZ dialogue system, by adding the capability to dialogue generation using the dataflow dialogue framework. 
% 
%By adding a user simulator (which simulates the behaviour of a human user) to an existing dialogue system (which already implements the agent side of the dialogue), we obtain a dialogue generator, which can generate both sides (user and agent) of a complete conversation.
%
% 
%In this work, we explore ways to create a user simulator for a dataflow dialogue system, and then use the user simulator to generate whole conversations. 

We explore agenda driven dialogue generation for the MultiWOZ domain, and a limited, agenda free user simulator for the SMCalFlow domain.

\section{Dialogue generation}

The term {\it dialogue generation} has previously been used for systems generating either one side (usually agent) or both sides (user and agent) of a conversation. In this paper, we concentrate on the latter case.

A module generating the user side of the conversation is referred to as a  {\it user simulator}, e.g. \cite{Schatzmann2009TheHA}. A major motivation for previous work on user simulators has been the high cost of obtaining human generated utterances.

 Especially in the development of dialogue systems, new user utterances are necessary after even a minor change in an automatic agent's response, as each dialogue turn can depend on all previous turns. In this case, the addition of a user simulator allows the developers of the dialogue system to continue  modifying and testing the system without the additional cost and delay brought on by the use of human users.
 
 A different perspective on the user simulator is to regard it as a test case generator, a common practice in software development, where tests are incrementally implemented in sync with new system functionalities.
 
In this work we explicitly distinguish between the natural language (unstructured) representation of a dialogue turn and its structured representation (in our case a dataflow expression). For the use of test case generation, we focus mainly on having the user simulator generate a structured representation, and use a simplistic template based method for generating the corresponding natural language user utterance. More varied and natural utterances could be generated e.g. following the methodology of \cite{fang2023truth}.

By running the (two way) dialogue generator and sampling examples of complete conversations, developers may be able to detect problems in the implementation of the agent part on dialogue paths they failed to consider (a common occurrence).

Another common use of dialogue generation is for creating a dataset to be used for training various dialogue models \cite{DBLP:journals/corr/abs-2107-11904} .
% \cite{TL-user-sim}

\section{Agenda based user simulator}

Agenda based user simulator (ABUS) has been previously used in cases where it's possible to specify a desired end state of a dialogue, which the user intends to reach. 

\cite{Schatzmann2009TheHA} described one such example for the MultiWOZ dataset, using a stack to push and pop user targets, and thus describe the dialogue state from the user's perspective, i.e. the user simulator drives the conversation so that it finally reaches the specified state (e.g. book a specific hotel with the prescribed date, duration, number of people etc).

In this work, we implemented a user simulator add-on to the dataflow implementation of MultiWOZ, with the goal of exploring and demonstrating the benefits and difficulties specific to the dataflow paradigm.

\subsection{MultiWOZ agenda}
As in other user simulator implementations for MultiWOZ, the agendas to be used by the dialogue generator are taken from  original  MultiWOZ dialogues.

However, rather than having a separate representation of the user agenda (e.g. using a separate stack), in the dataflow implementation the agenda is represented by a dataflow graph (with the same format as the one used for the execution of dialogues). This is possible since the dataflow graphs hold both functions and data.

The selected dialogue is first executed by the dataflow MultiWOZ system \cite{mwoz_df}, which results in a set of dataflow graphs corresponding to the progression of the selected dialogue.

\begin{figure*}
\includegraphics[height=0.26\textheight,trim= 0.0in 0.5in 0.5in 0.5in]{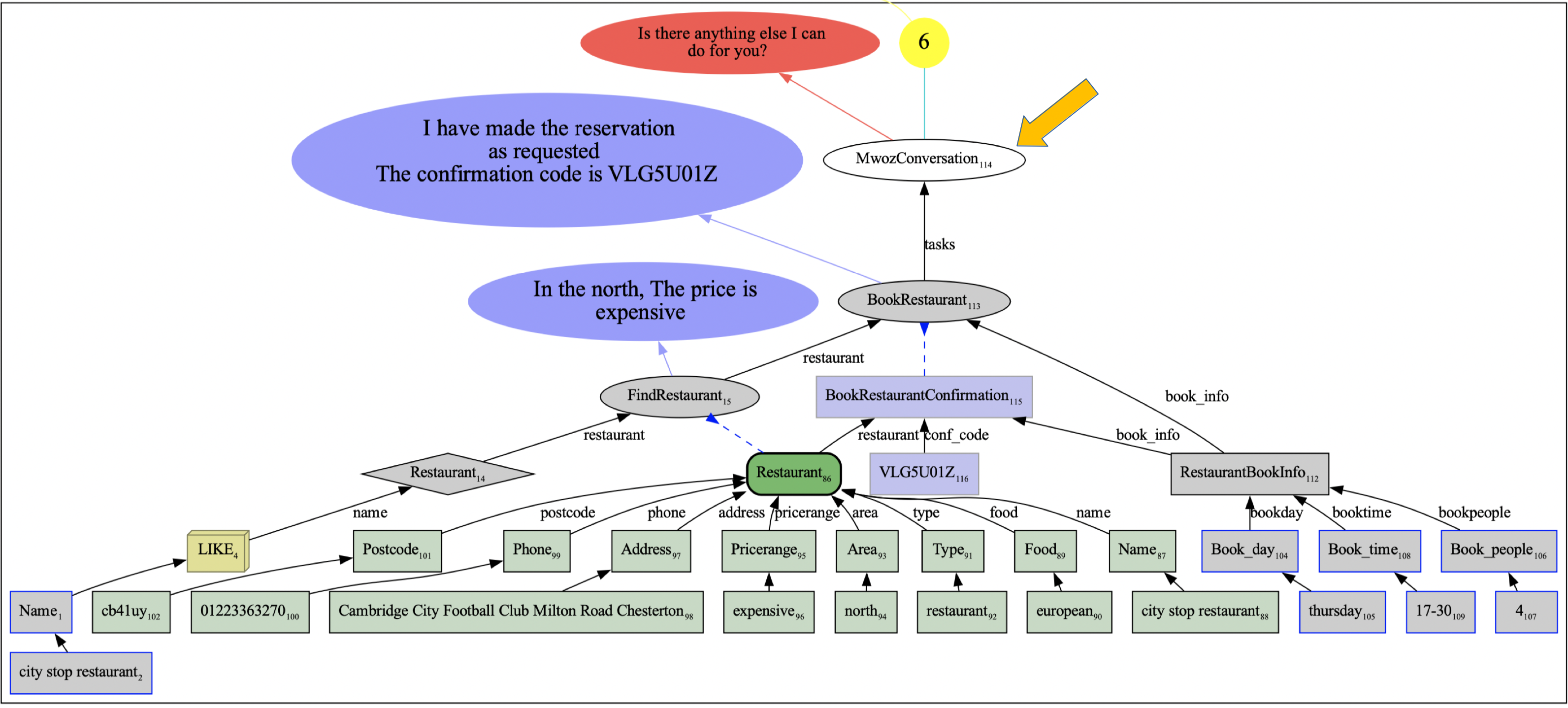}
\caption{\label{agenda} Dataflow graph under the last MwozConversation node (indicated by the orange arrow) for the execution of a short MultiWOZ dialogue. This is taken as the agenda to use by the user simulator. Black arrows indicate inputs, dashed blue arrows indicate result.}
\end{figure*}

We take the final dataflow graph (more precicely, the graph under the last MwozConversation node, as shown in the example in  figure \ref{agenda}).

This graph represents the final state of the dialogue (which may differ from the agenda given as a target to the human user when the dataset was created, in cases the dialogue failed to reach the prescribed target, e.g when the requested task was not possible to be fulfilled, or the users gave up), containing the final state of the user's request 
%(including, e.g.,  cases where the user changed their mind during the dialogue) 
and the information added by the agent (additional information found in the domain database as well as any bookings made by the agent).

We use this graph as our {\it target} graph. During the dialogue generation, we start from a new (empty) graph, and progressively modify it, trying to reach the same state (i.e. make the new graph be equivalent to the target graph).

%Note that this agenda may differ from the agenda declared in the MultiWOZ instructions for the same dialogue, as here we extract the agenda from the actual execution of the dialogue, which may not always reach the declared goals.

\section{Dataflow user simulator}
In the dataflow implementation of the MultiWOZ dialogue system \cite{mwoz_df}, the inputs (natural language) come from a human user, and are converted by a seq2seq translation model into dataflow expressions: {\it revise()} expressions are used to set slot values, and {\it refer()} is used to extract values which are not explicitly given in the current input. 

%These dataflow expressions are executed by the system (representing the agent's responses), modifying the dataflow graph representation of the dialogue. 

%The modified graph is evaluated by the agent logic, which either succeeds to fulfil the request, or else (when information is missing or wrong) notifies the user about the errors and/or makes suggestions.

For a dialogue generator, the dataflow expressions representing user requests are created programmatically by a user simulator, instead of being the result of translating natural language. 

Dialogue generation starts by initializing the new {\it current} dataflow graph (optionally generating some greeting message), which is then driven towards the {\it target} graph. 

This intuitively correlates to a mental model of natural dialogues, where information  is collected from both the user and agent utterances, and combined into a representation of the dialogue (including its history),

Concretely, in the dataflow dialogue paradigm, the dialogue representation is done using dataflow graphs, where graph nodes expect specific inputs (we refer to them here as "slots", even though they don't always match the definition of slots in typical intent/entity dialogue systems) of specific types, also allowing compositional structures.

At each successive turn, we expect the user's request to be "relevant" -  have some connection with the conversation  (its current state, or its history), i.e. it {\it can} be connected in some way to the existing graphs (practically, it is explicitly {\it allowed} to connect with the graphs). This connection could be simple (e.g. adding the value of a slot to the current task), or more complex (e.g. adding a value to a slot of a different task, or creating a new task)\footnote{It is entirely up to the developers of the dialogue system to decide what should be allowed to connect to the current graph.}.

For each further turn, the following steps are executed:

\subsection{Mapping}
First, we try to find all the nodes in the {\it current} graph which {\it could} be extended. This is done by finding the mapping between the current graph and the target graph, and for each {\it current} node which is successfully mapped to the target graph, custom logic is applied to decide if this node can be extended. 
For example, for the graph in figure \ref{agenda},  we allow extending the node RestaurantBookInfo (by setting a value to one of its inputs), but we do not allow to extend the terminal node "Thursday" (this is a design decision. A different approach is explored below for SMCalFlow).

\subsection{Node selection}
Once we have the set of nodes from the current graph which could be extended, we can select any of them.

In our implementation we randomly select one of these nodes, so separate runs of generation using the same target can lead to different dialogue paths.

\subsection{request selection}
Once a node has been selected to be extended, we call its custom logic for creating the user request.

In the basic case, this logic goes over the expected inputs of the selected ({\it current}) node, comparing them to the inputs of the mapped {\it target} node, looking for inputs which are still missing, or have different values. Then, one of this inputs is selected, and the generated user request specifies the value taken from the target node's input as the new value for this input for the current node.

Alternatively, more than one slot can be selected to be requested in one turn.

In order to simulate users mistakes, or users changing their mind, we can also allow (with some probability) the value used for the selected slot to differ from the value in the target graph (randomly select from a set of alternative values for this slot).

An additional source of slot values are values which were mentioned previously in the {\it current} dialogue. Once we have selected the value to be used in the user request, we can search the {\it current} graph to see if the value already appeared in it (using the {\it refer()} function), and if so, we can choose (with some probability) to use the reference instead of explicitly repeating the value.

\subsection{Request generation}
The selected request (as described above) is formulated as a dataflow expression (graph), whose execution will modify the dialogue graph according to the user's request.

In addition, a natural language description of the user's request is generated by rules. These rules can take advantage of the hierarchical structure of the graph, where each node needs only to know how to describe itself, and how to compose the natural language descriptions of its inputs into a coherent description.

As mentioned previously, this simplistic natural language generation could be improved by using more sophisticated approaches, e.g. \cite{fang2023truth}, but for many applications (e.g. generating use cases for debugging, or augmenting translation training datasets, as described below) this simple approach is sufficient. 

\section{Persona}

Similar to \cite{persona}, where the concept of a {\it Persona} is used to characterize different behaviours and traits of the {\it agent} side of the dialogue, we use this term to also apply to the (automated) {\it user} side of the dialogue, and have collected parameters controlling the creation of the user request into a Persona object. 

This controls whether the user is likely to:
\begin{itemize}
\item Request just one slot in a turn, or multiple ones
\item Make mistakes / change their mind
\item Use reference for slot values or not
\item Answer the agent's question (or respond to the agent's suggestion) or ignore it
\item Prematurely end the dialogue, before it reaches the target state.
\end{itemize}

\section{Using functions as slot values}

To further increase the diversity of the generated dialogues, and to better emulate natural user requests, the values used for inputs are allowed to be function calls, rather than just terminal values (integers, strings etc.). For example, use:
\begin{itemize}
\item "Two days after Monday" instead of "Wednesday"
\item "John's manager" instead of "Dan"
\end{itemize}

These functions need to return a value of the right type, as well as a value which is semantically appropriate.

This opens the door for composition of several functions, which is common in natural language.

Since the dataflow implementation of MultiWOZ is quite limited in the number of functions it implements (due to the simplicity of the domain), we implemented a dialogue generator for the SMCalFlow domain, which has many more functions, and which was explicitly designed with function composition in mind.

\section{Replacing values by functions}
The actual replacement of values by functions is done recursively as follows.

we start with a value {\it V} (of a specific type {\it T}), given as input {\it I} to some function {\it F}.

For example "Dan" is the value (of type {\it Person}), which is used as input to the {\it attendee} slot of {\it CreateEvent()}. 
In this case: {\it V}=Dan, {\it T}=Person, {\it I}=attendee, {\it F}=CreateEvent.

We first scan all the functions which are available in the SMCalFlow implementation. We use the typing information of the dataflow implementation to select only the functions which return the correct type ({\it T}).

For each of these  functions, we check if they are allowed/able to generate value {\it V} as the input {\it I} for the function {\it F}. 

This is implemented as custom logic, which should enforce coherent use of the functions, as well as prevent functions which are of the right type, but wrong value. 

For example,  {\it Yesterday()} returns a date, but it won't return a general target value. 

Another example would be when trying to replace a {\it Person} value by using the {\it GetManager()} function, but there is no matching instance in the corresponding knowledge base (e.g. if that person is not a manager of any employee). 

Note that the inclusion of these semantic constraints in constructing compositional user requests (e.g. dependence on external knowledge base, and other custom logic) may be difficult (or impossible) using standard CFG mechanisms.

The dataflow implementation uses a typing system which reflects the semantics of the domain. This is used during function replacement to ensure that  functions return a semantically compatible value. This prevents the generation of expressions like: {\it meeting.duration=GetTemperature(Zurich)}, which would be easy to execute, but probably not what a user is going to ask (and whose conversion to natural language request is awkward).

\section{Dialogue generation without agenda}
Combining the use of (compositional) functions as slot values with agenda driven dialogue generation,  implies that the values returned by the functions should match the value specified by the agenda. For example, if the agenda specifies that we should create a meeting with "Dan", and we choose to use the function {\it GetManager()} (which returns the manager of a given person), then we have to find an input to {\it GetManager()} which will return "Dan" (requiring a knowledge base containing the necessary objects and relations, such as employees, managers, etc.)

Unfortunately, Semantic Machines did not release the knowledge base needed to run the SMCalFlow dialogues (as well as the knowledge that would be needed for function composition for the generation task).

While manual creation of this knowledge for one dialogue (such that it is logically consistent with the given dialogue) is not difficult, we lacked the resources to manually do this at SMCalFlow scale.

Instead, for exploring dialogue generation with function composition for SMCalFlow, we relaxed the need for the functions to return the value specified by a dialogue agenda. All we require is that the value returned is of the right formal type, and that it is semantically allowed (as explained above). Note that in this implementation we generate only the first user turn of the dialogue (since we do not have the knowledge base required to support coherent completion of the dialogue).

Starting with a given dataflow graph (expression), we iterate over the  same steps as described for the MultiWOZ agenda driven dialogue generation, but adapted to no-agenda SMCalFlow domain (specifically, the custom logic of the SMCalFlow domain).

For all the "simple" (i.e. non-function) nodes in the graph, we check if they can be generated by a function, i.e. if there are any functions which can be used to replace them (as described in the previous section).

We then randomly select one of the nodes and one of the functions which can replace it (for example, if a Person node was selected to be replaced by a function, we randomly select between "GetManager", "GetFriend", "get  attendee of event",  etc.)

%For the actual replacement, we implement custom logic for each function to generate the actual expression to be used.

For function composition, we allow several iteration of this process, each time replacing a terminal value by a function call (e.g. "Dan" $ \rightarrow$ "GetManager(John)" $ \rightarrow$ "GetManager(GetFriend(Jill))" $ \rightarrow$ ...).

The final expression is converted to natural language using the same process as described for the MultiWOZ case.

The number of iterations (corresponding to the depth of composition) could be randomly selected, but it was observed that in practice depths of more than 3 often resulted in requests which were less natural - both the content (i.e. humans would rarely ask  such complex questions) and the natural language rendering of the request (e.g. humans would use case specific formulations to avoid ambiguities arising when describing nested functions).

\section{Augmenting training data}
 
It has been previously observed that machine learning (specifically sequence to sequence) models have difficulty with generalization when handling compositions, especially when lacking examples of specific combinations in the training data \cite{bogin}. 

To alleviate this problem, we try using the dialogue generator to generate examples covering all possible compositional patterns (up to a given depth), and use these to augment the training data of the seq2seq translation model.

To do this, we ran the dialogue generator for the SMCalFlow domain for 10 million iterations, generating variations of {\it CreateEvent()} expressions.

In each iteration, a random subset of the inputs of {\it CreateEvent()} is selected and initialized (randomly selecting a value from a list of allowed values), and then the values are replaced by functions as described previously.

We then filter the resulting expressions to keep only one example for each unique "structure" (we consider two dataflow expressions to have the same structure if they are an exact match after removing terminal values). This left 1.2 million examples, each one being a pair: [natural language, dataflow expression] (both generated by the dialogue simulator, as explained previously).

We used these examples to augment the original training set of SMCalFlow. Since the original training set is much smaller than the augmentation data (it has about 130K utterances), it was replicated (upsampled) 5 times and then mixed with the augmentation data. 

We then used the same pipeline which was used in the original SMCalFlow paper \cite{andreas2020task-oriented} to train and evaluate the translation seq2seq model. This pipeline uses OpenNMT \cite {opennmt}.

We also experimented training adapter layers for the Flan-T5 model \cite{hChung22}. Adapters are new layers that are inserted in between layers of a pre-trained model. The weights of these layers are then fine-tuned to the given task, while the remaining weights of the pre-trained model are frozen \cite{jPfeiffer20}. For this experiment we trained adapter layers for the Flan-T5 base model (248M parameters), using the default values of the adapter-transformers\footnote{\url{https://github.com/adapter-hub/adapter-transformers}} library \cite{jPfeiffer20}. We trained for $20$ epochs, with a learning rate of $2\times10^{-5}$	and batch size of $4$. 

Table \ref{aug_exp} shows that with this augmentation both training methods resulted in accuracy improvements. 

This improvement occurred  despite the fact that the natural language generated by the user simulator is template based and shows very little variation (as well as generating some grammatically incorrect sentences). An optimistic interpretation is that the model is able to learn from the original dataset how to deal with language variability, and from the generated augmentation data how to deal with compositional variations.

%As we can be seen from Table \ref{aug_exp}, the use of the augmented dataset brought an improvement of approximately $2.5\%$, which is similar to the improvement we observed on SMCalFlow's original pipeline.

\begin{table}
\centering
\begin{tabular}{ll|l}
\hline
 Training data & OpenNMT & flan-t5 \\
\hline
Original training set & 73.8\% &  75.3\%\\
\hline
Augmented training set & 76.2\% & 77.8\%\\
\hline
\end{tabular}
\caption{\label{aug_exp}
Accuracy of translating natural language to dataflow expression (validation set, exact match). Adding compositional augmentation data improves accuracy.}
\end{table}

\section{Conclusion}

In this work we presented the implementation of a user simulator in the dataflow paradigm, which, when added to an already existing dataflow dialogue system, can be used for generating both user and agent turns of a dialogue.

For the MultiWOZ domain, we have implemented an agenda based user simulator (ABUS), which can be used to generate complete and varied conversations given an agenda (target), by randomly sampling user actions from all the possible actions. We use a {\it Persona} to control the range of possible actions and the sampling strategy. 

In order to increase the variety of user requests, we have allowed the use of function compositions in the user simulator of the SMCalFlow domain. Due to lack of data and resources, we have implemented a  limited version of this user simulator, which does not use an agenda, and in effect generates only one turn of the dialogue (but with the necessary data, this can be extended to full dialogue generation).

While the concept behind the generation of our  dataflow expressions  is very close to CFG's, our implementation makes use of semantics and external knowledge bases (which can be dynamic), which would be a challenge for standard CFG's.

We have shown the output of the limited SMCalFlow can be used to augment the training dataset used for training the model which translates natural language user requests to dataflow expressions, which results in an increase in translation accuracy. We hypothesize that this improvement is the result of the added variety of function compositions in the examples generated by the user simulator.

The code for dialogue generation has been added to the OpenDF repository, which should allow replication and extension of the work described in this work.

\section{Bibliographical References}\label{reference}

\bibliographystyle{lrec2022-bib}
\bibliography{gen_df}

\appendix

\section{Example generated dialogues}\label{exExpr}

\subsection{MultiWOZ}
Tables \ref{gen_mwoz}  shows  and example dialogue generated by our dialogue generator for the MultiWOZ domain.

\begin{table*}
\centering
\begin{tabular}{llll}
\hline
Turn & Speaker & Text \\
\hline
1 & User & hello\\
1 & Agent & Hello, I'm your MWOZ agent. How can I help you?\\
2 & User & I'm looking for a restaurant on Thursday\\
2 & Agent & I see several (20) matches, maybe select name or address\\
3 & User & I'm looking for a restaurant at 13:00\\
3 & Agent & I see several (20) matches, maybe select name or address\\
4 & User & I'm looking for a restaurant called city stop restaurant\\
4 & Agent & OK, … For how many people would you like to book the restaurant?\\
5 & User & what is the address of the restaurant?\\
5 & Agent &For city stop restaurant  the address is ...  For how many people would you like to book the restaurant?\\
6 & User & I'm looking for a restaurant for 4 people, at 17:30\\
6 & Agent & I have made the reservation ...  Is there anything else I can do for you?\\
7 & User & Goodbye!\\
7 & Agent & Goodbye!\\

\hline
\end{tabular}
\caption{\label{gen_mwoz}
An example dialogue generated for the MultiWOZ domain. In this dialogue, the user simulator decided to emulate a user mistake, and gave the wrong time (13:00) at turn 3. At turn 5, the user simulator decided to ignore the agen't answer, and ask for information instead. At the end of the dialogue, the specified agenda has been satisfied. }
\end{table*}

\subsection{SMCalFlow}

Table \ref{gen_smcal} shows  example turns generated by the dialogue generator for the SMCalFlow domain. For each turn, a dataflow expression and its natural language rendering are shown (using the simplified SMCalFlow annotation format presented in \cite{simplifying}).

\begin{table*}
\centering
\begin{tabular}{lll}
\hline
 source & text \\
\hline
NL & create an event lasting 25 minutes at the location of the event next weekend \\
&at the location of the event with Dan and John\\

DF & $CreateEvent(AND(has\_duration(toMinutes(25)),$ \\
& $ at\_location(:location(FindEvents(AND(at\_location(:location(FindEvents($\\
&$AND( with\_attendee(Dan) , with\_attendee(John))))), starts\_at(NextWeekend( ))))))))$\\

\hline

NL & create an event the day of starting of the get together lasting 3 weeks with the friend of Adam\\

DF & $CreateEvent( starts\_at( NextDOW( :dow( :start( FindEvents( $\\
&$AND( with\_attendee( singleton( FindFriends( Adam ) ) ) , $\\
&$has\_duration( toWeeks( 3 ) ) , has\_subject( get together ) ) ) ) ) ) ) )$\\
\hline

NL & create an event lasting 1 week with the manager of the person who attended the meeting \\
&lasting 3 years next weekend\\

DF & $CreateEvent( AND( with\_attendee( FindManager( :recipient( :attendees( FindEvents( $\\
&$AND( has\_duration( toYears( 3 ) ) , starts\_at( NextWeekend( ) ) , $\\
&$has\_subject( meeting ) ) ) ) ) ) ) , has\_duration( toWeeks( 1 ) ) ) )$\\

\hline
NL & create a get together at the location of the event lasting 2 months at the time of \\
&starting of the event lasting 1 week today\\

DF & $CreateEvent( AND( at\_location( :location( FindEvents( AND( starts\_at( $\\
&$NumberAM( :time( :start( FindEvents( AND( has\_duration( toWeeks( 1 ) ) , $\\
&$starts\_at( Today( ) ) ) ) ) ) ) ) , has\_duration( toMonths( 2 ) ) ) ) ) ) , has\_subject( get \ together ) ) )$\\

\hline
\end{tabular}
\caption{\label{gen_smcal}
Example dataflow expressions and their corresponding natural language utterances generated by the dialogue generator for the SMCalFlow domain.  The natural language description of deep compositions can easily become ambiguous or unnatural.}
\end{table*}

\end{document}